\documentclass[letterpaper, 10 pt, conference]{ieeeconf}  % Comment this line out if you need a4paper

\IEEEoverridecommandlockouts                              % This command is only needed if 
\usepackage{amsmath} % assumes amsmath package installed
\usepackage{amssymb}  % assumes amsmath package installed
\usepackage{cite}
\usepackage{booktabs}
\usepackage[subnum]{cases}
\usepackage{mathtools}
\usepackage{balance}
\usepackage{color}
\usepackage{tabularx}
\usepackage{graphicx,dblfloatfix} 
\usepackage{listings}
\usepackage{textcomp}
\usepackage{url}
\usepackage{multirow}
\usepackage{algorithm}
\usepackage[noend]{algpseudocode}
\usepackage{ragged2e} 
\newcolumntype{Y}{>{\RaggedRight\arraybackslash}X} 

\usepackage{bm}
\usepackage{bbm}
\usepackage[colorinlistoftodos]{todonotes}
\usepackage{hyperref}
\usepackage[normalem]{ulem}
\usepackage{units}
\usepackage{threeparttable} 
\usepackage{pifont}% http://ctan.org/pkg/pifont
\usepackage[T1]{fontenc}
\usepackage{eucal}

\newcommand{\cmark}{\ding{51}}%
\newcommand{\xmark}{\ding{55}}%

\usepackage[printonlyused]{acronym}

\acrodef{RSL}{Robotic Systems Lab}
\acrodef{COM}{center of mass}
\acrodef{SQP}{sequential quadratic problem}
\acrodef{VMC}{virtual model controller}
\acrodef{w.r.t.}{with respect to}
\acrodef{DOF}{degrees of freedom}
\acrodef{ZMP}{zero moment point}
\acrodef{COP}{center of pressure}
\acrodef{IMU}{inertial measurement unit}
\acrodef{COT}{cost of transport}
\acrodef{JPL}{Jet Propulsion Laboratory}
\acrodef{HAA}{hip adduction/abduction}
\acrodef{HFE}{hip flexion/extension}
\acrodef{KFE}{knee flexion/extension}
\acrodef{ZMP}{zero-moment point}
\acrodef{QP}{quadratic programming}
\acrodef{SQP}{sequential quadratic programming}
\acrodef{WBC}{whole-body controller}
\acrodef{HO}{hierarchical optimization}
\acrodef{NLP}{nonlinear programming}
\acrodef{MPC}{model predictive control}
\acrodef{TO}{trajectory optimization}
\acrodef{DARPA}{Defense Advanced Research Projects Agency}
\acrodef{JPL}{Jet Propulsion Laboratory}
\acrodef{RBDL}{Rigid Body Dynamics Library}
\acrodef{RL}{reinforcement learning}
\acrodef{LIP}{linear inverted pendulum}
\acrodef{COP}{center of pressure}
\acrodef{SLQ}{sequential linear quadratic}
\acrodef{DDP}{differential dynamic programming}
\acrodef{SRBD}{single rigid body dynamics}
\acrodef{EOM}{equations of motion} % acronyms

 \newcommand{\deleted}[1]{}

\title{\LARGE \bf
Whole-Body MPC and Online Gait Sequence Generation for Wheeled-Legged Robots
}

\author{Marko Bjelonic, Ruben Grandia, Oliver Harley, Cla Galliard, Samuel Zimmermann and Marco Hutter
\thanks{
This work was supported in part by armasuisse W\&T and the Swiss National Science Foundation (SNF) through the National Centres of Competence in Research Robotics (NCCR Robotics) and Digital Fabrication (NCCR dfab). Besides, it has been conducted as part of ANYmal Research, a community to advance legged robotics.}% First sentence use only for final RAL version.
\thanks{All authors are with the Robotic Systems Lab, ETH Z\"urich, 8092 Z\"urich, Switzerland.
{\tt\footnotesize marko.bjelonic@mavt.ethz.ch}}
}

\begin{document}

% Paper headers 
%\markboth{IEEE Robotics and Automation Letters. Preprint Version. Accepted January, 2020}{Bjelonic \MakeLowercase{\textit{et al.}}: Rolling in the Deep -- Hybrid Locomotion for Wheeled-Legged Robots using Online Trajectory Optimization}
% Use only for final RAL version

\maketitle
% \pagestyle{empty}
% \thispagestyle{empty}
% Comment or remove these lines for final RAL version.

%%%%%%%%%%%%%%%%%%%%%%%%%%%%%%%%%%%%%%%%%%%%%%%%%%%%%%%%%%%%%%%%%%%%%%%%%%%%%%%%
\begin{abstract}
Our paper proposes a model predictive controller as a single-task formulation that simultaneously optimizes wheel and torso motions. This online joint velocity and ground reaction force optimization integrates a kinodynamic model of a wheeled quadrupedal robot. It defines the single rigid body dynamics along with the robot's kinematics while treating the wheels as moving ground contacts. With this approach, we can accurately capture the robot's rolling constraint and dynamics, enabling automatic discovery of hybrid maneuvers without needless motion heuristics. The formulation's generality through the simultaneous optimization over the robot's whole-body variables allows for a single set of parameters and makes online gait sequence adaptation possible. Aperiodic gait sequences are automatically found through kinematic leg utilities without the need for predefined contact and lift-off timings, reducing the cost of transport by up to \unit[85]{\%}. Our experiments demonstrate dynamic motions on a quadrupedal robot with non-steerable wheels in challenging indoor and outdoor environments. The paper's findings contribute to evaluating a decomposed, i.e., sequential optimization of wheel and torso motion, and single-task motion planner with a novel quantity, the prediction error, which describes how well a receding horizon planner can predict the robot’s future state. To this end, we report an improvement of up to \unit[71]{\%} using our proposed single-task approach, making fast locomotion feasible and revealing wheeled-legged robots' full potential.
\end{abstract}

%\begin{IEEEkeywords}
%Legged Robots, Wheeled Robots, Motion and Path Planning, Optimization and Optimal Control
%\end{IEEEkeywords} % Use only for final RAL version

%%%%%%%%%%%%%%%%%%%%%%%%%%%%%%%%%%%%%%%%%%%%%%%%%%%%%%%%%%%%%%%%%%%%%%%%%%%%%%%%
\section{INTRODUCTION}
%
%\IEEEPARstart{L}{egged} % Use only for final RAL version
Quadrupedal robots are fast becoming more common in industrial facilities~\cite{bellicoso2018jfr}, and it is only a matter of time until we see more of these robots in our daily lives. Their locomotion capabilities are well understood, and there are many different approaches published that exploit knowledge about their natural counterparts~\cite{eckert2015comparing,nyakatura2019reverse}. The understanding of locomotion principles has led to simplified models and heuristics that are widely used as templates to control legged robots~\cite{sardain2004forces,Kalakrishnan2010Fast,Rebula2007Controller,Focchi2020Heuristic,Zucker2011Optimization,kajita2001inv,Wieber2006Trajectory}. While legged robots have already made their way into real-world applications, wheeled-legged robots are still (mostly) only within the research community~\cite{klemm2019ascento,klamt2020remote,dietrich2016whole,Reid2020Mobility,cordes2018design,geilinger2020Computational}. Their locomotion capabilities are less understood due to missing studies of natural counterparts and the additional \ac{DOF} of the wheels, making simplified models that capture dynamic \emph{hybrid locomotion}, i.e., simultaneous walking and driving, cumbersome to design.
\begin{figure}[t]
    \centering
    \includegraphics[width=\columnwidth]{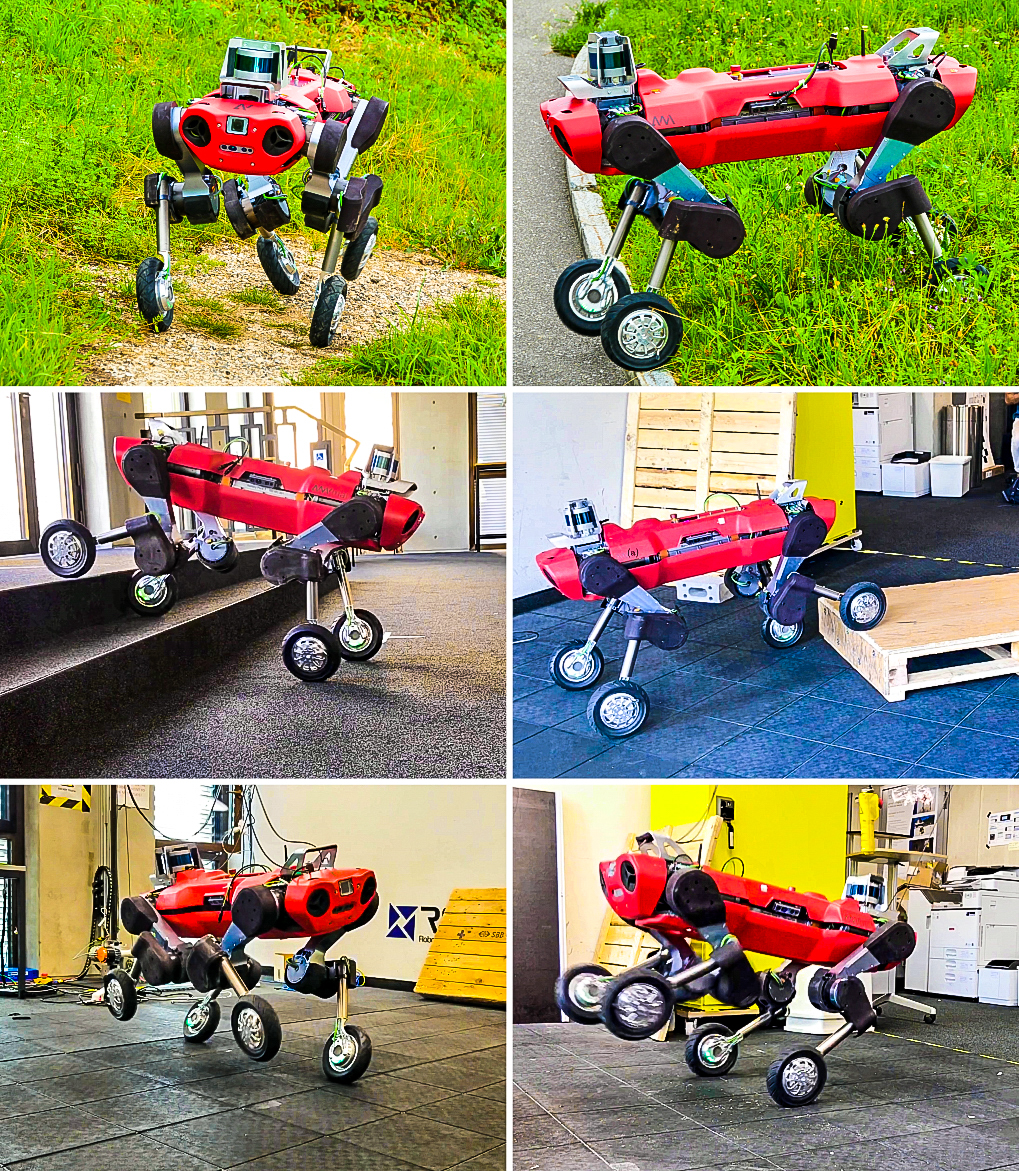}
    \caption{With our novel whole-body \ac{MPC}, the robot ANYmal~\cite{hutter2017anymaljournal}, equipped with actuated wheels, explores indoor and outdoor environments in a fast and versatile way (video available at \href{https://youtu.be/_rPvKlvyw2w}{https://youtu.be/\_rPvKlvyw2w}). First row: Locomotion in high grass and over steep hills of up to \unit[2]{m/s}, while gait sequences are automatically discovered. Second row: Blindly stepping over a \unit[0.20]{m} high step (\unit[32]{\%} of leg length) and stairs with a \unit[0.175]{m} high step (\unit[28]{\%} of leg length). Third row: Pacing gait and \unit[0.28]{m} high jump with front legs.}
    \label{fig:anymal_on_wheels}
\end{figure}

Hybrid locomotion for robots, such as depicted in Fig.~\ref{fig:anymal_on_wheels}, faces two specific problems, one requires \emph{continuous}, and the other \emph{discrete} decision-making. The latter relates to the task of finding the appropriate gait sequencing, i.e., sequences of lift-off and touch-down timings, which becomes difficult to handcraft. Besides, the work in~\cite{bjelonic2020rolling} reveals that the proper choice of gait sequences for wheeled-legged robots is crucial to reducing the \ac{COT}. The former problem describes the task of finding the continuous motion of the robot, i.e., the trajectories of the torso and wheels. Our \emph{whole-body}\footnote{In this paper, the \ac{MPC}'s whole-body model includes the whole-body kinematics and single rigid body dynamics, simultaneously optimizing the robot's contact forces, generalized coordinates and velocities, including the 6D torso motion and joint kinematics.}
\ac{MPC} requires minimal assumptions about the robot's dynamics and kinematics, allowing wheeled-legged robots to accurately capture the rolling constraint without adding unnecessary assumptions. 
\subsection{Related Work}
In the following sections, we categorize existing approaches to legged locomotion and bring them into the context of hybrid locomotion.
\subsubsection{Continuous Decision-Making}
\label{sec:intro:tasks}
A \emph{decomposed-task approach} splits the problem into separate foot (or wheel) and torso tasks. By breaking down locomotion planning for high-dimensional \mbox{(wheeled-)legged} robots into two lower-dimensional sub-tasks, we hypothesize that the individual problems become more tractable. The coordination of each task's solution is one of the main challenges, and heuristics are needed to align the foot and torso motions. Many approaches were developed over the last years exploiting these task synergies~\cite{kajita2001inv,Wieber2006Trajectory,Rebula2007Controller,Kalakrishnan2010Fast,Zucker2011Optimization,engelsberger2014overview,Kudruss2015Optimal,Park2015Online,Wieber2016Modeling,Naveau2017Reactive,jenelten2020Perceptive,buchanan2020perceptive}.

In contrast, a \emph{single-task approach} treats the continuous decision problem as a whole without breaking down the problem into several sub-tasks~\cite{Erez2013integrated,Dai2014Whole,farshidian2017Efficient,Caron2017When,orin2013centroidal,kuindersma2016optimization}. Here, the challenge is to solve the problem in a reasonable time, so that online execution on the real robot becomes feasible. In the last few years, traditional legged locomotion research experienced a large amount of pioneering work in the field of \ac{MPC}~\cite{Erez2013integrated,Dai2014Whole,farshidian2017Efficient,Caron2017When,Koenemann2015wholebody} that now reliably runs on quadrupedal robots, like \emph{ANYmal}~\cite{grandia2019feedback,Neunert2018WholeBody}, and \emph{MIT Cheetah}~\cite{bledt2019implementing}. Another class of single-task optimization problems involves \ac{TO} that precomputes complex trajectories over a time horizon offline~\cite{kuindersma2016optimization,Mordatch2012Discovery,Herzog2015Trajectory,winkler2018gait}. Hybrid locomotion platforms, e.g., \emph{Skaterbots}~\cite{geilinger2020Computational}, \emph{RoboSimian}~\cite{bellegarda2019trajectory} and walking excavators~\cite{jelavic2019wholebody}, provide a similar approach to motion planning over flat terrain by solving a \ac{NLP} problem.

The dynamic model and underlying foothold heuristic are two essential aspects of continuous decision-making:
\paragraph*{Dynamic Models}
Optimization-based methods depend on the choice of model complexity. Each dynamic model comes with its assumptions. For example, the \ac{LIP} model controls only the motion of the \ac{COM} position and acts as a substitute for the contact forces. Here, the \ac{ZMP}~\cite{vukobratovic2004zero} is constrained to lie inside the support polygon~\cite{sardain2004forces,Kalakrishnan2010Fast,Kalakrishnan2011Learning,Zucker2011Optimization,caron2017zmp,viragh2019trajectory,bellegarda2018Design,Mastalli2020Motion}. These approaches result in fast update rates at the cost of inaccurate modeling of the real robot.

The real model can be approximated more accurately with a \ac{SRBD} model, which assumes that the joint accelerations' momentum is negligible and that the full system's inertia remains similar to some nominal configuration. Recent years showed impressive results, and many different research groups have adopted this more complex model~\cite{Herzog2015Trajectory,farshidian2017Efficient,Caron2017When,grandia2019feedback,winkler2018gait,bledt2019implementing,di2018dynamic}.

Finally, the rigid body dynamics model only assumes non-deformable links, and the \ac{EOM} can be rewritten as the Centroidal dynamics model~\cite{orin2013centroidal,kuindersma2016optimization,Budhiraja2019Consensus}. Such a dynamic model is common in \ac{TO} and provides a general approach to hybrid locomotion~\cite{geilinger2020Computational}. Due to the increased complexity, these hybrid motions are impractical to update online with feedback control.
\paragraph*{Foothold Heuristics}
As described in Section~\ref{sec:intro:tasks}, a decomposed-task approach is completed in two stages, where a heuristic is needed to connect the feet and torso planning stages. For example, a common method in legged locomotion designs foothold positions based on the Raibert heuristic~\cite{raibert1986legged} with a capture-point-based feedback term~\cite{pratt2006capture}. The work in~\cite{bledt2019implementing} regularizes a single-task \ac{MPC} using such kinds of heuristics, which might guide the optimization problem towards sub-optimal solutions due to the heuristic's simplicity. In our previous work, this approach is also referred to as inverted pendulum models~\cite{bjelonic2020rolling}. Its design is not intuitive for hybrid locomotion since it assumes a single foothold.
\subsubsection{Discrete Decision-Making}
Gaits in legged robots are often hand-tuned and time-based. Moreover, appropriate sequences of contact timings become hard to design when it comes to wheeled-legged robots, as shown in Fig.~\ref{fig:anymal_on_wheels}.

Including discrete decision variables into the continuous decision-making results in a holistic approach, as shown by~\cite{aceituno_mastalli17ral,deits2014footstep,winkler2018gait,Mordatch2012Discovery}. These approaches achieve impressive results, but their algorithms are currently impractical to run online on the real robot in a feedback control loop. Finding gait sequences in a separate task might reduce the problem's complexity and make online execution on the robot feasible. By considering the impulses that the legs can deliver, online gait adaptation is shown by the \emph{MIT Cheetah} robot~\cite{Boussema2019Online}. The authors, however, reduce the problem to 2D due to the computational complexity of the 3D case and split the continuous motion planning into decomposed tasks.
\subsection{Contribution}
We extend the related work with a \emph{whole-body \ac{MPC}} allowing for \emph{online gait sequence adaptation}. The former finds the robot's torso and wheels' motion in a single task by introducing a novel kinodynamic model of a wheeled-legged robot that incorporates the wheels as moving ground contacts with a fixed joint position and an accurate estimation of the rolling constraint. Moreover, the \ac{MPC} optimizes the joint velocity and ground reaction force simultaneously and allows for a single set of parameters for all hybrid motions, which enables us to adapt the sequences of contact and swing timings. In short, our main contributions are:

\mbox{\textbf{1) Hybrid Locomotion.}} We evaluate whole-body \ac{MPC} for a wheeled-legged robot, providing a single-task approach that automatically discover complex and dynamic motions that are impossible to find with a decomposed-task approach. Due to the kinodynamic model, our framework accurately captures the real robot's rolling constraint and dynamics.

\mbox{\textbf{2) Comparison.}} We compare the performance of a decomposed- and single-task approach on the same robotic platform. In this regard, we introduce a quantity that allows us to compare different motion planning algorithms through the \emph{prediction accuracy}, which describes how well a receding horizon planner can predict the robot's future state.
    
\mbox{\textbf{3) Discrete Decisions.}} Our \ac{MPC} performs all behaviors with the same set of parameters, enabling flexibility regarding the gait sequence and allowing us to propose a concept to quantify kinematic leg utilities for online gait sequence generation without the need for predefined contact timings and lift-off sequences. This automatic gait discovery lets wheeled quadrupedal robots, as depicted in Fig.~\ref{fig:anymal_on_wheels}, to coordinate aperiodic behavior and reduce the \ac{COT} drastically.
\section{PROBLEM FORMULATION}
\label{sec:mpc}
The general \ac{MPC} formulation is to find the control input of the following optimization over a receding horizon $T$ based on the latest state measurement $\bm{x}_0$. Its optimized control policy is applied to the robot at each iteration until an updated policy is available.
\begin{subequations}
\label{eq:mpc}
\begin{align}
& \underset{\bm{\bm{u}(\mathord{\cdot})}}{\text{minimize}}
& & \phi(\bm{x}(T)) + \int_{0}^{T} l(\bm{x}(t),\bm{u}(t),t) \mathrm{d}t, \label{eq:mpc_cost}\\
& \text{subjected to}
& & \dot{\bm{x}}(t) = \bm{f}(\bm{x}(t),\bm{u}(t),t), \label{eq:system_dynamics}\\
& & &\bm{x}(0) = \bm{x}_0, \label{eq:initial_contidition}\\
& & &\bm{g}_1(\bm{x}(t),\bm{u}(t),t) = 0, \label{eq:state_input_equality_constraint}\\
& & &\bm{g}_2(\bm{x}(t),t) = 0, \label{eq:state_equality_constraint}\\
& & &\bm{h}(\bm{x}(t),\bm{u}(t),t) \geq 0. \label{eq:inequality_constraint}
\end{align}
\end{subequations}
where $\bm{x}(t)$ is the state vector and $\bm{u}(t)$ is the control input vector at time $t$. Here, $l(\mathord{\cdot})$ is the time-varying running cost, and $\phi(\mathord{\cdot})$ is the cost at the terminal state $\bm{x}(T)$. The state-input equality constraint (\ref{eq:state_input_equality_constraint}), pure state equality constraint (\ref{eq:state_equality_constraint}), and inequality constraint (\ref{eq:inequality_constraint}) are handled by a Lagrangian method, penalty method, and relaxed barrier function, respectively. Our \ac{MPC} formulation relies on the \ac{SLQ} approach of~\cite{farshidian2017Efficient} with the feedback policy of~\cite{grandia2019feedback}, which is a \ac{DDP}~\cite{mayne1966second} based algorithm for continuous-time systems.

\begin{figure}[t]
    \centering
    \includegraphics[width=\columnwidth]{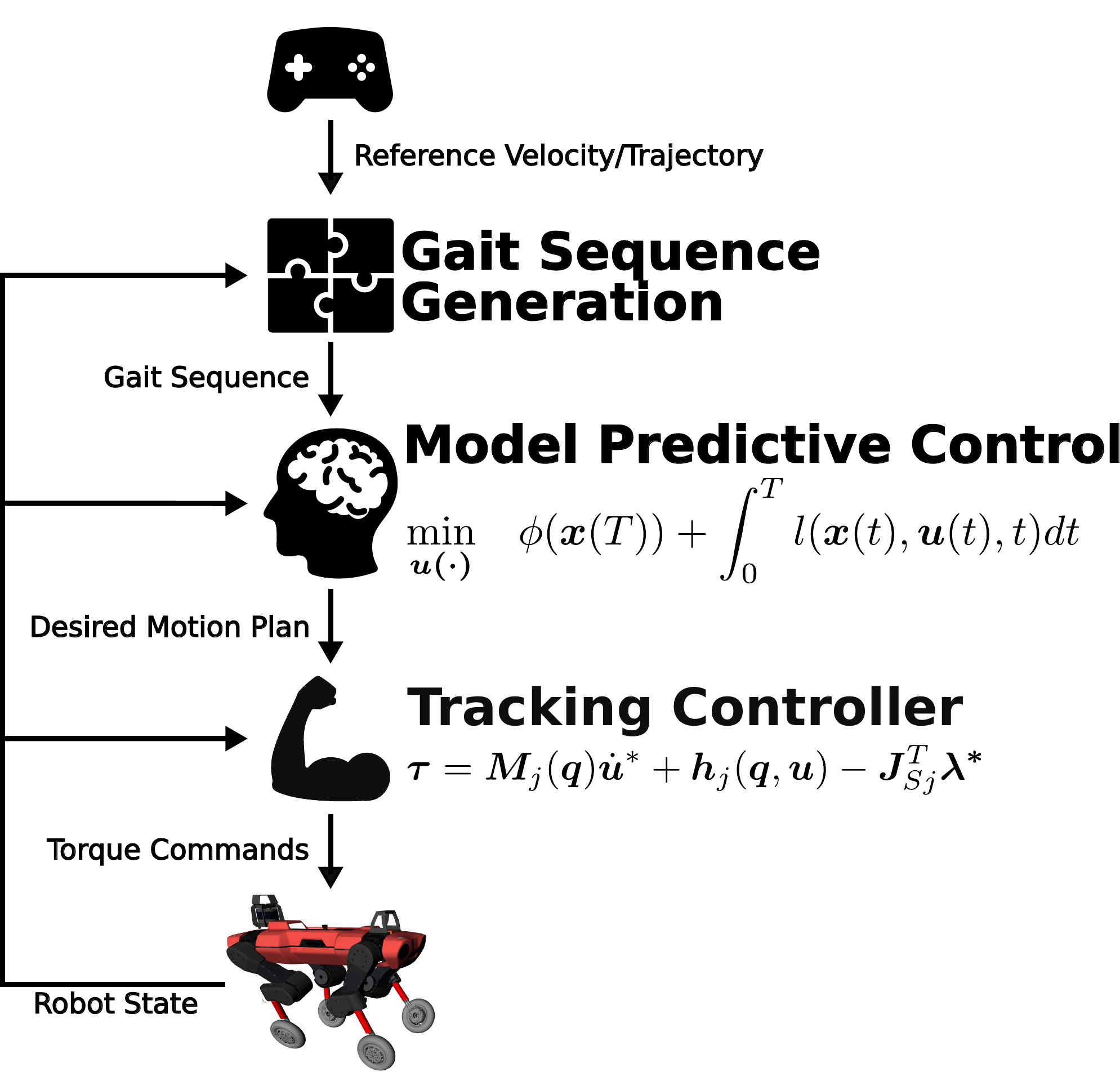}
    \caption{Overview of the locomotion controller. The gait sequence generator automatically transforms reference trajectories from a higher-level planner or operator device into lift-off and touch-down sequences. These gait sequences are fed into the \ac{MPC} that optimizes joint velocities and contact forces over a time horizon $T$. Finally, a tracking controller, e.g., \cite{bjelonic2019keep}, transforms the desired motion plan into torque references $\bm{\tau}$.}
    \label{fig:overview}
\end{figure}

Fig.~\ref{fig:overview} visualizes our complete locomotion controller that is verified in challenging experiments at the end of this paper. In the following, we introduce our main contributions, the \ac{MPC}'s implementation and online gait sequence generation for wheeled-legged robots, in more detail. 
\section{HYBRID LOCOMOTION}
The remainder of this section proposes a solution for hybrid locomotion, as depicted in Fig.~\ref{fig:anymal_on_wheels}, which is challenging due to the additional motion along the rolling direction, making the design of motion primitives and gait sequences impossible to hand-tune.
\subsection{Model Predictive Control Implementation}
In this work, we avoid motion primitives by proposing a single-task \ac{MPC} optimizing over the robot's whole-body variables. We continue with the underlying wheeled-legged robot's model, and the \ac{MPC}'s cost function and constraints.
\subsubsection{Modeling}
\begin{figure}[t]
    \centering
    \includegraphics[width=\columnwidth]{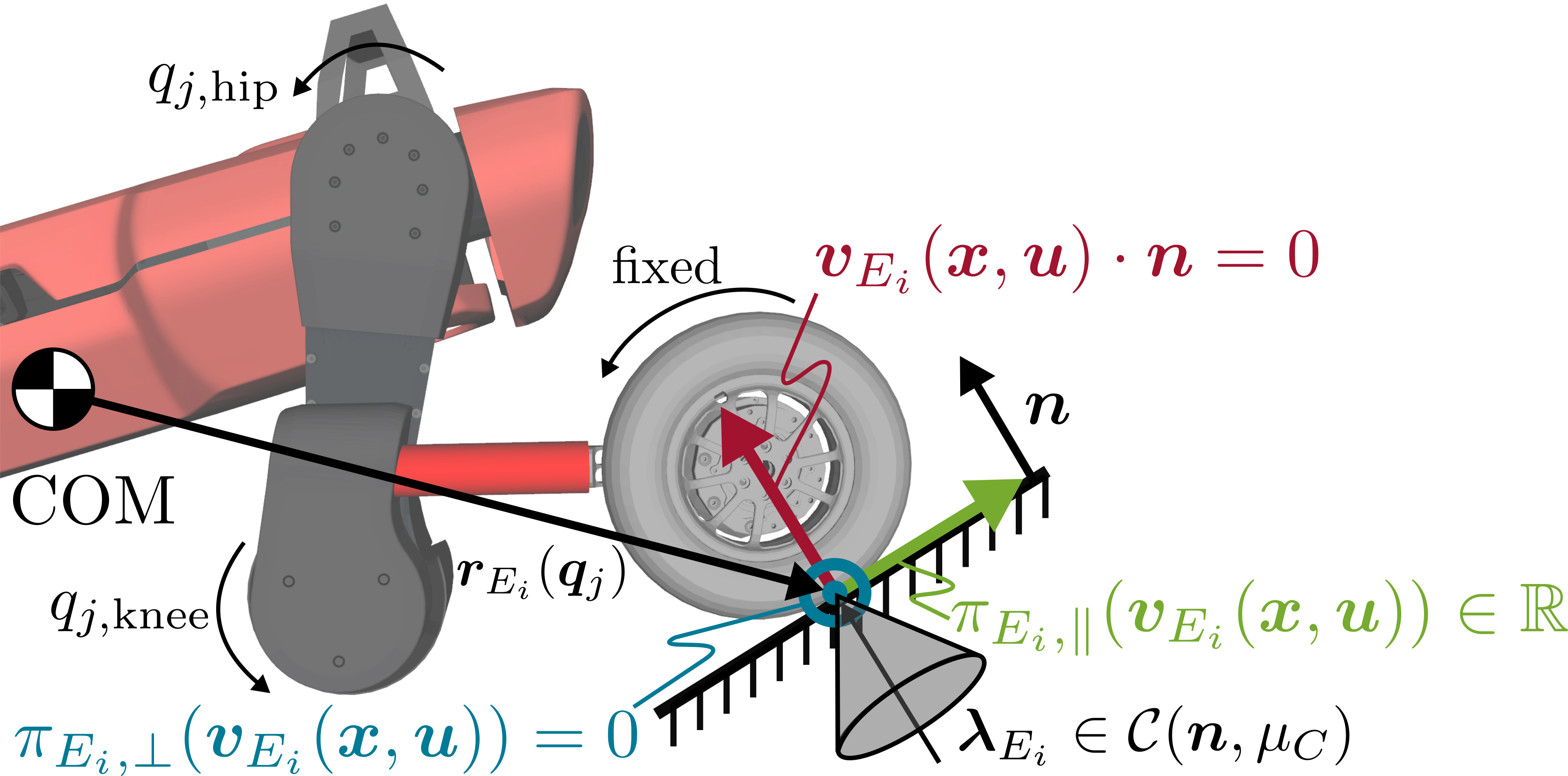}
    \caption{Sketch of the rolling constraint with the underlying wheel model as a moving point contact with a fixed joint position. The image shows each direction of the end-effector velocity $\bm{v}_{E_i}(\bm{x},\bm{u})$, end-effector contact position $\bm{r}_{E_i}(\bm{q}_j)$, and friction cone constraint $\bm{\lambda}_{E_i} \in \CMcal{C}(\bm{n},\mu_C)$.}
    \label{fig:rolling_constraint}
\end{figure}
Adding a full model of a wheel increases the \ac{MPC}'s number of states and inputs $n$ by two per leg, which increases the optimization time since the \ac{SLQ}'s backward pass scales by $(n)^3$. As shown in Fig.~\ref{fig:rolling_constraint}, we model the robot's wheel as a moving point contact with a fixed joint position, which can be translated into wheel inputs through the wheel's contact velocity and radius. With this novel formulation, the \ac{MPC}'s optimization time does not increase compared to legged robots~\cite{grandia2019feedback} despite the additional \ac{DOF}.

We let the frame $E_i$ be fixed at a leg’s endpoint, i.e., the point on the wheel that is in contact with the ground during stance phase, and define this point as a leg's end-effector. This enables us to model conventional point-foot and wheels by changing the kinematic constraints and avoids additional constraints concerning the wheel. The state vector $\bm{x}(t)$ and control input vector $\bm{u}(t)$ in (\ref{eq:mpc}) are
\begin{subequations}
\label{eq:opt_variables}
\begin{align}
\bm{x}(t) &= 
\begin{bmatrix}
\bm{\theta}^T & \bm{p}^T & \bm{\omega}^T & \bm{v}^T & \bm{q}_j^T
\end{bmatrix}^T \in \mathbb{R}^{12 + n_j}, 
\label{eq:state_vector}\\
\bm{u}(t) &= 
\begin{bmatrix}
\bm{\lambda}_E^T & \bm{u}_j^T
\end{bmatrix}^T \in \mathbb{R}^{3n_e + n_j},
\label{eq:control_input_vector}
\end{align}
\end{subequations}
where $n_j=12$ and $n_e=4$ are the number of joints (excluding the wheel) and legs. The elements $\bm{\theta}$, $\bm{p}$, $\bm{\omega}$, $\bm{v}$ and $\bm{q}_j$ of the state vector in (\ref{eq:state_vector}) refer to the torso's orientation in Euler angles, torso's position in world frame $W$, \ac{COM}'s angular rate, \ac{COM}'s linear velocity, and joint positions, respectively. Moreover, the control inputs in (\ref{eq:control_input_vector}) are the end-effector contact forces $\bm{\lambda}_E$ and joint velocities $\bm{u}_j$.
\subsubsection{Cost Function}
We are interested in following external commands fed into a quadratic cost function of the state and control input vector. Thus, the time-varying running cost in (\ref{eq:mpc_cost}) is given by
\begin{equation}
\label{04:eq:mpc_cost_implemented}
\begin{aligned}
l(\bm{x}(t),\bm{u}(t),t) = \frac{1}{2} \Tilde{\bm{x}}(t)^T \bm{Q} \Tilde{\bm{x}}(t) + \frac{1}{2} \Tilde{\bm{u}}(t)^T \bm{R} \Tilde{\bm{u}}(t),
\end{aligned}
\end{equation}
where $\bm{Q}$ is a positive semi-definite Hessian of the state vector error $\Tilde{\bm{x}}(t)=\bm{x}(t)-\bm{x}_\mathrm{ref}(t)$ and $\bm{R}$ is a positive definite Hessian of the control input vector error $\Tilde{\bm{u}}(t)=\bm{u}(t)-\bm{u}_\mathrm{ref}(t)$. The error vector require reference values for the whole-body, e.g., the torso's reference position and linear velocity are computed through an external reference trajectory\footnote{The reference trajectory is generated from an external source, e.g., an operator device or a navigation planner. In case of reference velocities, i.e., the linear $\bm{v}_{\mathrm{ref}}$ and angular velocity vector $\bm{\omega}_{\mathrm{ref}}$ of the \ac{COM}, the reference trajectory can be computed by integrating these velocities.} $\bm{r}_{B,\mathrm{ref}}(t)$ of the torso $B$. The remaining variables of $\bm{x}(t)$ and $\bm{u}(t)$ are regularized to some nominal configuration.
%its derivative, and reference orientation and angular velocity is identical and zero, respectively. The joint reference states are set to some nominal position while the joint velocity vector is regularized to zero. Finally, the reference contact forces are simply calculated by dividing the robot's total mass by the number of contact legs.
%
\subsubsection{Equations of Motion}
The system's dynamics (\ref{eq:system_dynamics}) is based on a kinodynamic model of a wheeled quadrupedal robot. It defines the \ac{SRBD} model along with the kinematics for each leg while treating the wheels as moving ground contacts with a locked rotational angle. \ac{SRBD} assumes that the limb joints' momentum is negligible compared with the lumped \ac{COM} inertia and the inertia of the full-body system stays the same as to some nominal joint configuration. The \ac{EOM} of the \ac{SRBD} is given by
\begin{subequations}
\label{eq:srbd}
\begin{align}
\dot{\bm{\theta}} &= \bm{T}(\bm{\theta})\bm{\omega}, \label{eq:srgb:euler_rates}\\
\dot{\bm{p}} &= \bm{R}_{WB}(\bm{\theta})\bm{v}, \label{eq:srgb:position}\\
\dot{\bm{\omega}} &= \bm{I}^{-1} \left( -\bm{\omega} \times \bm{I} \bm{\omega} + 
\sum_{i=1}^{n_e} \bm{r}_{E_i}(\bm{q}_j) \times \bm{\lambda}_{E_i} \right), \label{eq:srgb:angular_acceleration}\\
\dot{\bm{v}} &= \bm{g}(\bm{\theta}) + \frac{1}{m} \sum_{i=1}^{n_e} \bm{\lambda}_{E_i}, \label{eq:srgb:linear_acceleration}\\
\dot{\bm{q}}_j &= \bm{u}_j, \label{eq:joint_velocity}
\end{align}
\end{subequations}
where $\bm{R}_{WB}(\bm{\theta}) \in SO(3)$ represents the rotation matrix that projects the components of a vector from the torso frame $B$ to the world frame $W$, $\bm{T}(\bm{\theta})$ is the transformation matrix from angular velocities in the torso frame $B$ to the Euler angles derivatives in the world frame $W$, $\bm{I}$ is the moment of inertia of the \ac{COM} taken at the nominal configuration of the robot, $m$ is the total mass, $\bm{g}(\bm{\theta})$ is the gravitational acceleration in torso frame $B$, and $\bm{r}_{E_i}(\bm{q}_j)$ is the end-effector's contact position of leg $i$ \ac{w.r.t.} the \ac{COM} (see Fig.~\ref{fig:rolling_constraint}), which is a function of the joint positions and thus, the kinodynamic model requires (\ref{eq:joint_velocity}).
\subsubsection{Rolling Constraint}
\label{sec:rolling_constraint}
The contact constraint of traditional legged robots is modeled through the end-effectors' velocities, and when in contact, these velocities are restricted to zero in all directions. Wheeled-legged robots, on the other hand, can execute motions along the rolling direction when in contact. Thus, the end-effector constraint of leg $i$ in contact is represented by
\begin{subequations}
\label{eq:rolling_constraint_contact}
\begin{align}
\bm{\lambda}_{E_i} &\in \CMcal{C}(\bm{n},\mu_C), \label{eq:rolling_constraint_contact:force}\\
\pi_{E_i, \bot}(\bm{v}_{E_i}(\bm{x},\bm{u})) &= 0, \label{eq:rolling_constraint_contact:rolling_side}\\
\bm{v}_{E_i}(\bm{x},\bm{u}) \cdot \bm{n} &= 0, \label{eq:rolling_constraint_contact:rolling_up}
\end{align}
\end{subequations}
where $\CMcal{C}(\bm{n},\mu_C)$ and $\bm{n}$ are the friction cone with its friction coefficient $\mu_C$  visualized in Fig.~\ref{fig:rolling_constraint} and the local surface normal in world frame $W$, respectively. The rolling constraint's sketch in Fig.~\ref{fig:rolling_constraint} shows each direction of the end-effector velocity $\bm{v}_{E_i}(\bm{x},\bm{u})$. Due to the kinodynamic model, the projection $\pi_{E_i, \bot}(\mathord{\cdot})$ in (\ref{eq:rolling_constraint_contact:rolling_side}) of the end-effector velocity in world frame $\bm{v}_{E_i}(\bm{x},\bm{u})$ onto the perpendicular direction of the rolling direction can be easily computed through forward kinematics. With this formulation, legs in contact are constrained, such that, the velocity along the rolling direction is left unconstrained, i.e, $\pi_{E_i, \parallel}(\bm{v}_{E_i}(\bm{x},\bm{u})) \in \mathbb{R}$. In contrast to \ac{SRBD} models without the robot's kinematics, our approach can accurately estimate the rolling constraint without introducing needless heuristics for its direction.

While leg $i$ is in air, the constraint switches to
\begin{subequations}
\label{eq:rolling_constraint_air}
\begin{align}
\bm{\lambda}_{E_i} &= \bm{0}, \\
\bm{v}_{E_i}(\bm{x},\bm{u}) \cdot \bm{n} &= c(t),
\end{align}
\end{subequations}
where legs in the air follow a predefined swing trajectory $c(t)$ in the direction of the terrain normal $\bm{n}$ and the ground reaction forces $\bm{\lambda}_{E_i}$ are set to zero.
%
%\subsubsection{Friction Cone Constraint}
%
%As given in (\ref{eq:rolling_constraint_contact}), the friction cone constraint $\bm{\lambda}_{E_i} \in \CMcal{C}(\bm{n},\mu_C)$ implements an inequality constraint which is given by
%
%\begin{equation}
%\label{eq:friction_cone}
%\begin{aligned}
%\mu_C F_z - \sqrt{F_x^2+F_y^2} \geq 0,
%\end{aligned}
%\end{equation}
%
%where the local contact force $\bm{F}=\begin{bmatrix}F_x & F_y & F_z \end{bmatrix}^T$ is a projection of the ground reaction forces $\bm{\lambda}_{E_i}$ to the frame of the terrain surface given by $\bm{n}$, and $\mu_C$ is the friction coefficient.
%
\subsection{Gait Sequence Generation}
\label{sec:wheelTrajectoryOptimization}
The \ac{MPC}'s implementation as a single task enables a single set of parameters for all motions, which further allows for the adaptation of its lift-off and touch-down timings. In this work, we exemplary show the implementation of a gait timings generation for multimodal robots with non-steerable wheels. Gait timings and their sequences are discovered through a kinematic utility of each leg. Given the external reference trajectory $\bm{r}_{B,\mathrm{ref}}(t)$, aperiodic sequences of contact and lift-off timings are generated over a time horizon $T$. 
%
%\subsection{Wheel Trajectory}
%\label{sec:reference_trajectory}
%
%The reference trajectory $\bm{r}_{B,\mathrm{ref}}(t)$ of the torso $B$ is given by integrating the velocity reference through
%
%\begin{equation}
%\label{eq:reference}
%\bm{r}_{B,\mathrm{ref}}(t) = \bm{r}_{B} + \int\limits_{0}^{t} \bm{R}(\theta_{\mathrm{yaw}} + \tau \bm{\omega}_{\mathrm{ref}}) \bm{v}_{\mathrm{ref}} \mathrm{d}\tau,
%\end{equation}
%
%whereby the rotation matrix $\bm{R}(\theta_{\mathrm{yaw}} + \tau \bm{\omega}_{\mathrm{ref}})$ describes the change in the torso's orientation caused by the reference yaw rate $\bm{\omega}_{\mathrm{ref}}$, $\bm{r}_{B}$ is the measured position of the \ac{COM}, and $\theta_{\mathrm{yaw}}$ is the measured yaw angle of the torso.
%
\subsubsection{Kinematic Leg Utility}
\label{sec:leg_utility}
For the robot to locomote, i.e., drive or walk, it needs to have a sense of each leg's utility $u_i(t) \in [0,1]$. If the utility of one leg approaches zero, the leg needs to be recovered by a swing phase. In contrast to~\cite{Boussema2019Online}, where the utility is based on impulse generation capabilities and used as a metric for a decomposed-task approach, we propose that the kinematic capability is of primary importance for gait adaptation of a single-task approach. This utility quantifies the usefulness of a leg in terms of remaining in kinematic reach.

Wheeled quadrupedal robots with non-steerable wheels, as shown in Fig.~\ref{fig:anymal_on_wheels}, have a fixed rolling direction. While in contact, the trajectory of the wheel $\bm{r}_{E_i,\mathrm{ref}}(t)=\bm{r}_{E_i}+\pi_{E_i, \parallel}(\bm{r}_{B,\mathrm{ref}}(t))$ is kinematically constrained, where $\bm{r}_{E_i}$ is the measured end-effector position of wheel $i$, and the projections $\pi_{E_i, \parallel}(\mathord{\cdot})$ and $\pi_{E_i, \bot}(\mathord{\cdot})$ are introduced in Section~\ref{sec:rolling_constraint}.

By defining the utility as an ellipse, we can distinguish the decay along and lateral to the rolling direction. Therefore, the leg’s utility $u_i(t)\in [0,1]$ is defined as
\begin{equation}
\label{eq:utility}
u_i(t) = 1 - \sqrt{
\left( \frac{\pi_{E_i, \parallel}(\widetilde{\bm{r}}_{E_i}(t))}{\lambda_{\parallel}} \right)^2+
\left( \frac{\pi_{E_i, \bot}(\widetilde{\bm{r}}_{E_i}(t))}{\lambda_{\bot}}\right)^2
},
\end{equation}
where the position error is given by $\widetilde{\bm{r}}_{E_i}(t)=\bm{r}_{B,\mathrm{ref}}(t)+\bm{r}_{BD_i}-\bm{r}_{E_i,\mathrm{ref}}(t)$, and $\bm{r}_{BD_i}$ is the position from the torso $B$ to the recent contact position at touch-down $D_i$ of leg $i$. $\lambda_{\parallel}$ and $\lambda_{\bot}$ are the two half-axis lengths of the ellipse along and lateral to the rolling direction and depend on the leg's kinematic reach.
\subsubsection{Gait Timings Generation}
The leg remains in contact as long as its utility $u_i(t)$ remains above a certain threshold $\bar{u} \in [0,1]$. If a leg's utility falls below the threshold, i.e., the leg is close to its workspace limits, then this leg is recovered by a swing phase with constant swing duration. Similar to~\cite{Boussema2019Online}, a multi-layered swing generator is proposed to achieve meaningful leg coordination:
\begin{enumerate}
    \item \label{04:utility_generation} \mbox{\textbf{Utility Generation.}} Calculate the utility for all legs $u_i(t)$ over a time horizon $T$.
    
    \item \label{04:utility_check} \mbox{\textbf{Utility Check.}} Find the time $t^*$ when $u_i(t)<\bar{u}$ and give legs with the lowest utility priority to add a swing phase with constant swing duration at time $t^*$.

    \item \label{04:neighboring_legs_check} \mbox{\textbf{Neighboring Legs Check.}} A swing phase is added if the neighboring legs\footnote{For the quadruped's left-front leg, the neighboring legs are the right-front and left-hind legs.} are not swinging. Otherwise, the swing phase is postponed until the neighboring legs are in contact—such an approach constrains the gaits to pure driving, hybrid static, and hybrid trotting gaits.
\end{enumerate}  
\section{EXPERIMENTAL RESULTS AND DISCUSSION}
\label{sec:experiments}
We validate our whole-body \ac{MPC} and gait sequence generation in several real-world experiments where we compare our approach's performance with the motion planner introduced in~\cite{bjelonic2020rolling}. It is based on a decomposed-task approach, i.e., the wheel and torso trajectories are solved sequentially. To the best of our knowledge, this is the first time a study compares the performance of a single- and decomposed-task approach on the same robotic platform. Table~\ref{table:capabilities} gives an overview of both approaches and lists their capabilities. Each element is described in more detail in the following sections, which reports on experiments conducted with ANYmal equipped with non-steerable, torque-controlled wheels (see Fig.~\ref{fig:anymal_on_wheels}). A video\footnote{\hbox{Available at \href{https://youtu.be/_rPvKlvyw2w}{https://youtu.be/\_rPvKlvyw2w}}} showing the results accompanies this paper.
\begin{table}[t!]
    \caption{Capabilities~\cite{winkler2018optimization} of our presented whole-body \ac{MPC}.}
    \label{table:capabilities}
    \begin{center}
    \begin{tabular}{ccc}
    \toprule
     &Whole-Body \ac{MPC} & Decomp. Task\cite{bjelonic2020rolling}\\\midrule
    \begin{tabular}{@{}c@{}}Dynamic model\\ (accuracy)\end{tabular} & Kinodynamic model &ZMP model \\[6pt]
    \begin{tabular}{@{}c@{}}Number of \\ optimizations\end{tabular} & Single optimization & \begin{tabular}{@{}c@{}}Separate wheel and \\ torso optimization \end{tabular} \\[6pt]
    \begin{tabular}{@{}c@{}}Foothold \\ heuristic\end{tabular} & No heuristics & \begin{tabular}{@{}c@{}}Inverted pendulum \\ model \end{tabular} \\[6pt]
    Update rate & \unit[20-50]{Hz} &\unit[100-200]{Hz}\\[6pt]
    Reliability & High &Medium \\[6pt]
    \begin{tabular}{@{}c@{}}Maximum reliable \\ speed\end{tabular} & \unit[2.5]{m/s} & \unit[1.5]{m/s} \\[6pt]
    Accelerations & High & Low \\ \\
    
    \multicolumn{3}{c}{\textbf{Optimized components}} \\
    Torso motion & {\color{green}6D}  & {\color{green}3D}\\
    Footholds & {\color{green}3D}  & {\color{green}2D} \\
    Swing leg motion & {\color{green}\cmark}  & {\color{green}\cmark} \\
    Contact force & {\color{green}\cmark}  & {\color{red}\xmark} \\
    Step timing/sequence & {\color{red}\xmark}  & {\color{red}\xmark} \\ \\
    
    \multicolumn{3}{c}{\textbf{Difficulty of shown task}} \\
    Line and point contacts & {\color{green}\cmark} & {\color{green}\cmark} \\
    Flight phases & {\color{green}\cmark}  & {\color{green}\cmark} \\
    Inclined terrain & {\color{green}\cmark}  & {\color{green}\cmark} \\
    Non-flat terrain & {\color{green}\cmark}  & {\color{red}\xmark} \\
    \begin{tabular}{@{}c@{}}Step timing/sequence \\ adaptation\end{tabular} & {\color{green}\cmark} & {\color{red}\xmark}\\
    \bottomrule
    \end{tabular}
    \end{center}
\end{table}
\subsection{Experimental Setup}
Our hybrid locomotion planner, tracking controller~\cite{bjelonic2019keep}, and state estimator~\cite{bjelonic2020rolling}, including the terrain normal estimation, run in concurrent threads on a single PC (Intel i7-8850H, 2.6 GHz, Hexa-core 64-bit). The robot is entirely self-contained in computation, and all optimization problems are run online due to fast solver times.
\subsection{Prediction Error}
Quantitatively comparing receding horizon planners based on the real robot's performance is a non-trivial task. In most cases, our community reports merely on the optimization time, success rate, and task difficulty without measuring its performance compared to other algorithms. Our work provides a novel quantity that describes how well a receding horizon planner can predict the robot's future state.

The optimization problem's ability to accurately predict the robot's state over a predefined time horizon is crucial for these planning algorithms. Measuring how accurately the underlying algorithm captures the real system is crucial. Therefore, we define the prediction error $\Delta p_{\mathrm{pred}}$ as
\begin{equation} \label{eq:prediction_error}
\Delta p_{\mathrm{pred}} = \lVert \bm{p}^{*}_{-T}(T) - \bm{p}_{\mathrm{meas}} \rVert, \hspace*{2mm} \forall \bm{v}_{\mathrm{ref}},\bm{\omega}_{\mathrm{ref}} = \mathrm{const.},  
\end{equation}
where $\bm{p}^{*}_{-T}(T)$ is the predicted \ac{COM} position, i.e., its terminal position optimized \unit[$T$]{s} ago, and $\bm{p}_{\mathrm{meas}}$ is the measured position of the \ac{COM}. Moreover, the prediction error is only computed for constant reference velocities $\bm{v}_{\mathrm{ref}}$ and $\bm{\omega}_{\mathrm{ref}}$.
\subsection{Decomposed- vs Single-Task Motion Planning}
In the following, we use a fixed trotting gait and compare the two approaches' performance in terms of their prediction error, dynamic model, and foothold heuristic.
\subsubsection{Prediction Accuracy}
\begin{figure}[t]
    \centering
    \includegraphics[width=\columnwidth]{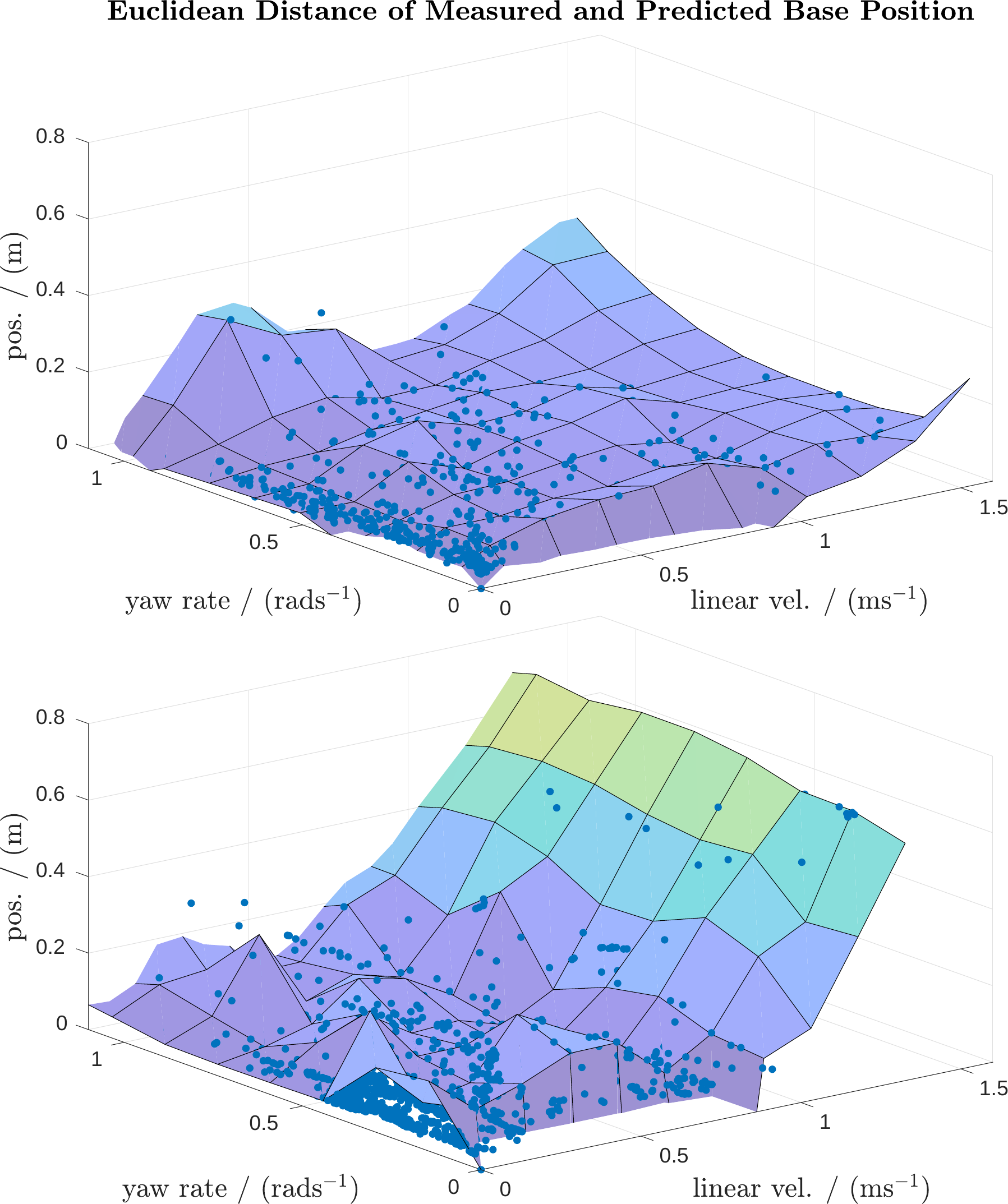}
    \caption{Prediction error for $T=0.8\mathrm{s}$ of the \ac{COM} while hybrid trotting on flat terrain. The upper figure depicts the result of our proposed whole-body \ac{MPC}, and the lower figure shows the result of our previously published decomposed-task approach~\cite{bjelonic2020rolling}. With our new locomotion controller, we achieve a prediction error of $\Delta p_{\mathrm{pred}}=0.061\pm0.044 \,\mathrm{m}$, which outperforms the result of the decomposed-task approach with $\Delta p_{\mathrm{pred}}=0.214\pm0.061\,\mathrm{m}$. Our single-task approach improves the prediction accuracy by approximately \unit[71]{\%}, which becomes evident at higher commanded linear velocities and yaw rates.}
    \label{fig:prediction}
\end{figure}

Fig.~\ref{fig:prediction} compares the performance of our whole-body \ac{MPC} with the decomposed-task approach described in~\cite{bjelonic2020rolling}. Especially at higher commanded velocities, the prediction error of the \ac{MPC} outperforms the prediction accuracy of our previously published controller, which is also prone to failures at higher speeds. Decoupling the locomotion problem into a wheel and torso task makes it untrackable at higher speeds. The actual wheel and torso trajectories start to diverge and require an additional heuristic to maintain balance. Our single-task approach, however, solves this problem and improves the prediction accuracy by up to \unit[71]{\%}, making fast locomotion feasible.
\subsubsection{Dynamic Model}
\label{sec:result:dynamic_model}
Various approaches use a \ac{LIP} model that optimizes over the \ac{ZMP} as a substitute for the contact forces. These approaches generate trajectories of the \ac{COM} so that the \ac{ZMP} lies inside the support polygon spanned by the legs in contact. The question arises whether this approach accurately captures the real dynamics. Therefore, we log the \ac{ZMP} of~\cite{bjelonic2020rolling} while running our \ac{MPC} using a more realistic kinodynamic model of a wheeled-legged robot.

\begin{figure}[t]
    \centering
    \includegraphics[width=\columnwidth]{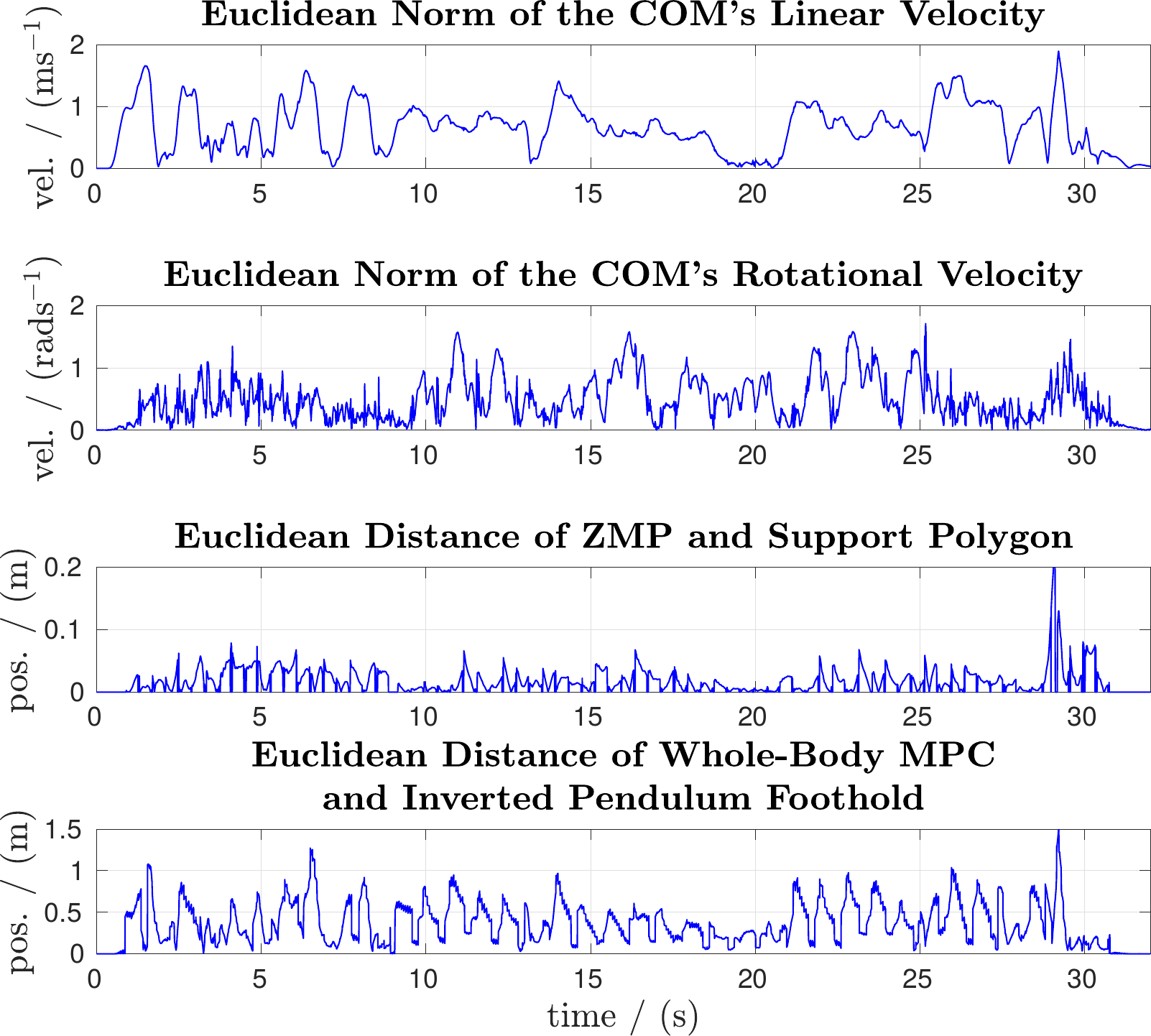}
    \caption{Results of our whole-body \ac{MPC} while commanding high torso accelerations, as shown in Fig.~\ref{fig:high_speed}. The upper two figures show the plot of the commanded linear and rotational velocities. As shown in the third plot, these motions are not feasible with a \ac{ZMP} model since the \ac{ZMP} lies outside the support polygon, i.e., the robot is supposed to fall. Similarly, the inverted pendulum model's heuristic in the last plot starts diverging from our approach's complex behaviors.}
    \label{fig:zmp_comparison}
\end{figure}

\begin{figure}[t]
    \centering
    \includegraphics[width=\columnwidth]{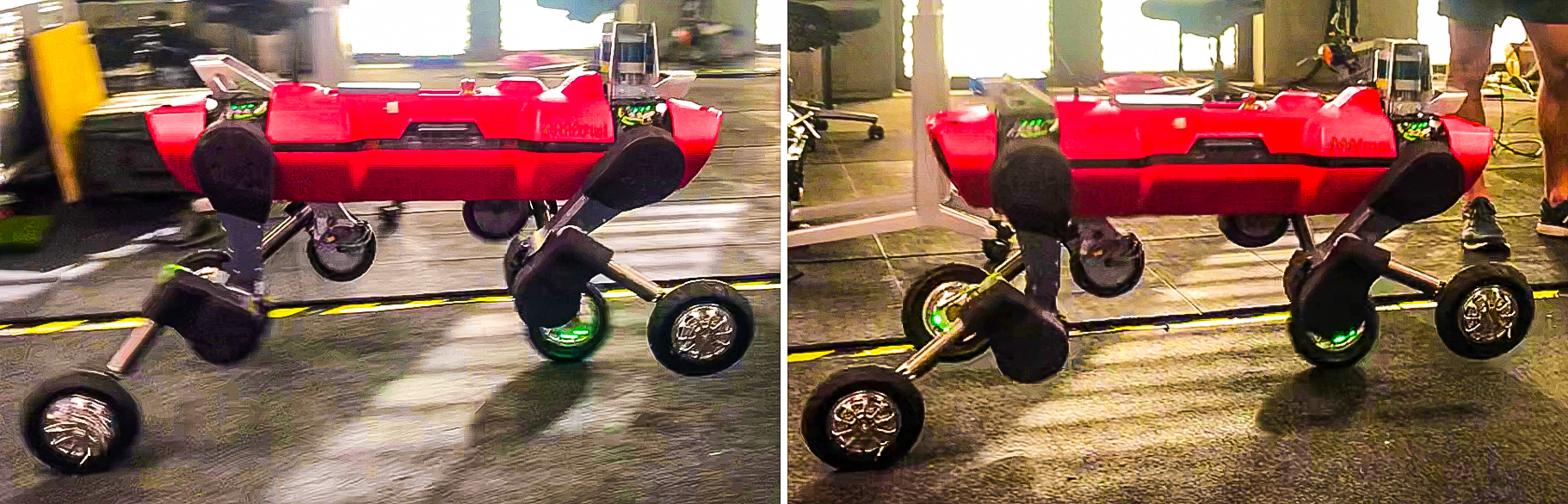}
    \caption{High accelerations using the whole-body \ac{MPC} approach. The robot executes a fast change of direction between 2 and \unit[-2]{m/s}, which forces the optimization problem to find complex motions that can not be captured by the \ac{LIP}, as shown in Fig.~\ref{fig:zmp_comparison}.}
    \label{fig:high_speed}
\end{figure}

The result in Fig.~\ref{fig:zmp_comparison} shows that while executing dynamic motions, the \ac{ZMP} diverges from the support polygon. Therefore, this simplified model can not discover motions as depicted in Fig.~\ref{fig:high_speed}. Furthermore, the idea of the \ac{ZMP} only holds in the presence of co-planar contacts~\cite{Orsolino2020Feasible}. Therefore, it can not accurately capture environments, as shown in the second row of Fig.~\ref{fig:anymal_on_wheels}, and thus, we need a more accurate model like the kinodynamic model presented here.
\subsubsection{Foothold Heuristic}
While the whole-body \ac{MPC} approach does not integrate any foothold heuristic, the decomposed-task approach relies on the inverted pendulum model based on a feedforward and feedback part. The former aligns the motions with the reference trajectory assuming a constant velocity of the torso. Simultaneously, the latter corrects the foothold under different conditions, such as modeling errors, external disturbances, and transitions. Similar to the result in Section~\ref{sec:result:dynamic_model}, Fig.~\ref{fig:zmp_comparison} shows that the inverted pendulum model diverges from our optimized footholds at higher accelerations due to the assumption of a constant reference velocity of the torso, which is tried to be compensated through the feedback term. Moreover, the inverted pendulum model adapts to unforeseen disturbances while stepping and is originally not designed for wheeled-legged robots. Handcrafting a heuristic as shown in~\cite{bjelonic2020rolling} that finds more dynamic and hybrid trajectories on the ground is cumbersome. Our approach discovers complex behaviors automatically (see Fig.~\ref{fig:high_speed}) thanks to the single-task approach.
\subsection{Gait Sequence Generation}
\begin{figure}[t]
    \centering
    \includegraphics[width=\columnwidth]{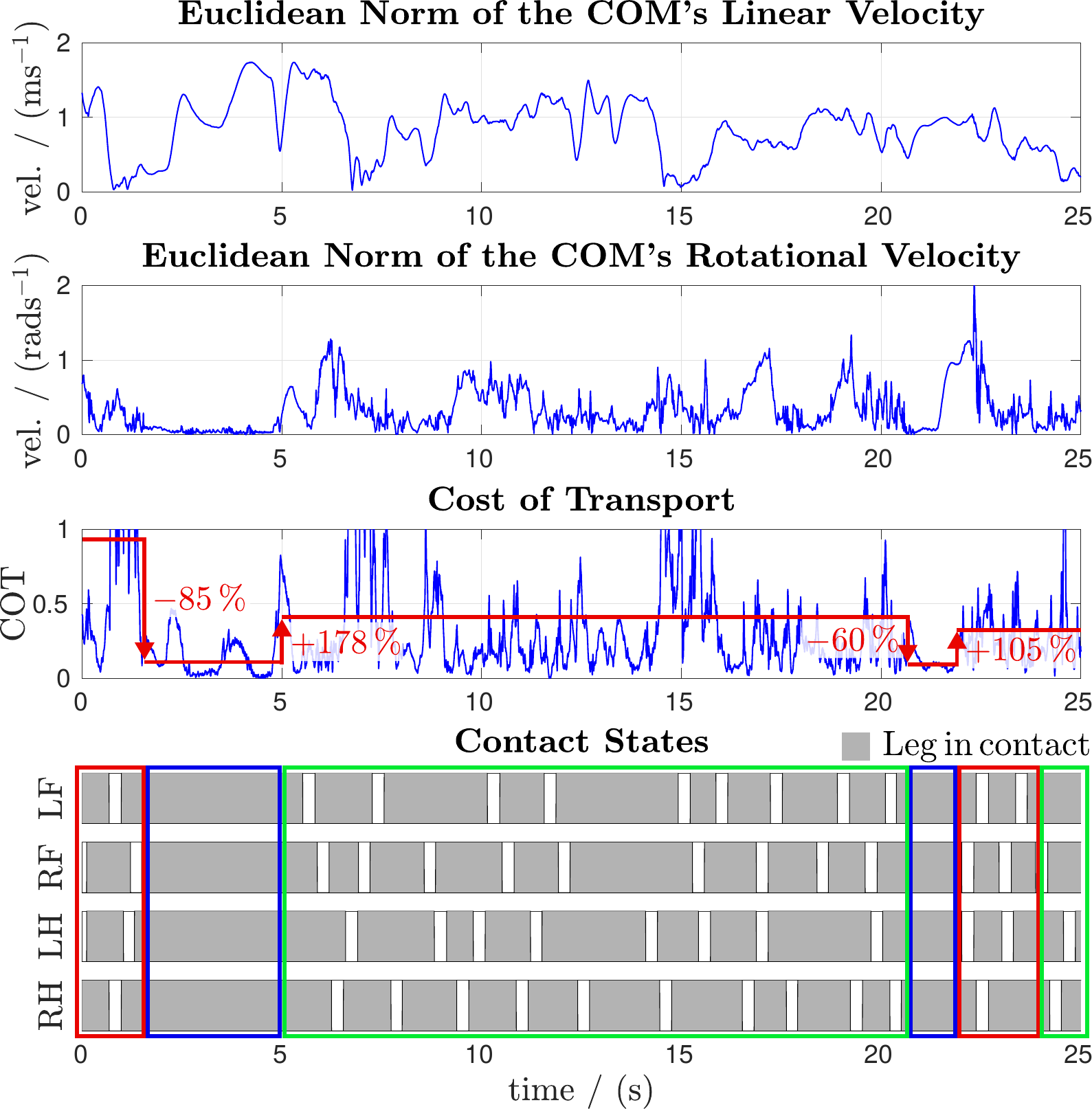}
    
    \vspace{0.2cm}
    
    \includegraphics[width=\columnwidth]{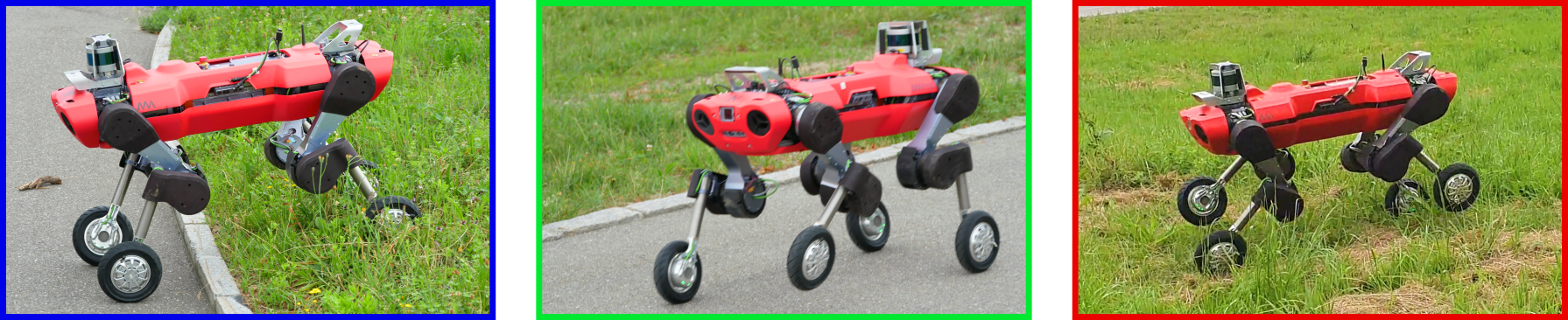}
    \caption{Contact timings diagram while running the gait sequence generator and whole-body \ac{MPC}. The two upper plots show the linear and rotational velocity of the \ac{COM}, the third plot depicts the mechanical \ac{COT}~\cite{bjelonic2018skating} including its average values, and the corresponding contact states are displayed in the four lower rows (left-front (LF), right-front (RF), left-hind (LH), and right-hind leg (RH)). The robot performs three different motions at high linear velocities in combination with no (\unit[1-5]{s}), medium (\unit[5-22]{s}), and high rotational velocities (\unit[22-23]{s}). As shown in the lower images, the gait sequence generator results in pure driving (blue box), hybrid static gaits (green box), i.e., one leg at a time, and hybrid trotting gaits (red box), respectively. Especially the pure driving phases reduce the \ac{COT} drastically.}
    \label{fig:gaitTimings}
\end{figure}

Fig.~\ref{fig:gaitTimings} shows the result of the gait sequence generation in combination with the whole-body \ac{MPC}. The plot shows three time snippets where the robot executes high linear velocities in combination with no, medium, and high rotational velocities. The gait sequence generator based on kinematic leg utilities intuitively switches between pure driving, static gaits (three legs in contact), and a trotting gait. As can be seen in the third plot of Fig.~\ref{fig:gaitTimings}, we can lower the \ac{COT} by up to \unit[85]{\%} thanks to the reduced number of steps. Moreover, pure driving achieves a \ac{COT} of around 0.1 at \unit[2]{m/s}, which is a factor of two higher than hybrid trotting \cite{bjelonic2020rolling}.

One of our \ac{MPC}'s benefits is that it uses one set of cost terms for each gait. By contrast, the decomposed-task approach, as described in~\cite{bjelonic2020rolling}, requires re-tuning the cost terms for each gait pattern. Therefore, it is not feasible to run our gait timings generator with such an approach without adding more heuristics that interpolate between sets of pre-tuned cost terms.
\section{CONCLUSIONS}
\label{sec:conclusions}
We present a novel whole-body \ac{MPC} for hybrid locomotion allowing for online gait sequence adaptation. It finds the robot's torso and wheels motion in a single task, where joint velocity and ground reaction forces are simultaneously optimized based on a kinodynamic model with moving ground contacts. The experimental results verify that our approach improves the model's accuracy and enables the robot to automatically discover hybrid and dynamic motions that are impossible to find through motion templates. Due to the single set of parameters, the \ac{MPC} is flexible \ac{w.r.t.} the gait sequence. Therefore, we integrate an online gait sequence generation based on kinematic leg utilities that makes predefined contact and swing timings obsolete. Our wheeled-legged robot ANYmal is now, for the first time, capable of coordinating aperiodic behavior, which decreases the overall \ac{COT} of our missions. In future work, we plan to further extend our (blind) gait sequence generation by augmenting its utility function with terrain information from exteroceptive sensors.
%


%%%%%%%%%%%%%%%%%%%%%%%%%%%%%%%%%%%%%%%%%%%%%%%%%%%%%%%%%%%%%%%%%%%%%%%%%%%%%%%%

%%%%%%%%%%%%%%%%%%%%%%%%%%%%%%%%%%%%%%%%%%%%%%%%%%%%%%%%%%%%%%%%%%%%%%%%%%%%%%%%

%%%%%%%%%%%%%%%%%%%%%%%%%%%%%%%%%%%%%%%%%%%%%%%%%%%%%%%%%%%%%%%%%%%%%%%%%%%%%%%%

%\section*{ACKNOWLEDGMENT}

%%%%%%%%%%%%%%%%%%%%%%%%%%%%%%%%%%%%%%%%%%%%%%%%%%%%%%%%%%%%%%%%%%%%%%%%%%%%%%%%

\balance
\bibliographystyle{IEEEtran}
\bibliography{IEEEabrv,submissionbibfile}

\end{document}